\documentclass[10pt,conference]{IEEEtran}
\usepackage{graphicx} 
\usepackage{booktabs}
\usepackage[table,dvipsnames]{xcolor}
\usepackage{hyperref}
\usepackage{amsfonts}
\usepackage{amsmath}
\usepackage{amssymb}
\usepackage{makecell}
\usepackage{multirow}

\newcommand{\E}{\mathbb{E}}
\newcommand{\Err}{\mathcal{E}}

\title{
Investigating the Robustness of Subtask Distillation under Spurious Correlation
}

\author{
\IEEEauthorblockN{
    Pattarawat Chormai\textsuperscript{1,2},
    Klaus-Robert M\"uller\textsuperscript{1,3,4,5}, and
    Gr\'egoire Montavon\textsuperscript{6,3,*}
    \thanks{*Corresponding author: gregoire.montavon@charite.de}
}
\IEEEauthorblockA{
    \textsuperscript{1}Machine Learning Group, Technische Universit\"{a}t Berlin (TU Berlin), 10587 Berlin, Germany \\
    \textsuperscript{2}Konrad Zuse School of Excellence in Learning and Intelligent Systems (ELIZA), 64289 Darmstadt, Germany \\
    \textsuperscript{3}BIFOLD\,--\,Berlin Institute for the Foundations of Learning and Data, 10587 Berlin, Germany \\
    \textsuperscript{4}Department of Artificial Intelligence, Korea University, Seoul 136-713, Korea \\
    \textsuperscript{5}Max Planck Institute for Informatics, 66123 Saarbr{\"u}cken, Germany  \\
    \textsuperscript{6}Institute for AI in Medicine, Charit\'e\,--\,Universit\"atsmedizin Berlin, 10115 Berlin, Germany
}
}

\date{January 2026}
\begin{document}

\maketitle

\begin{abstract}
Subtask distillation is an emerging paradigm in which compact, specialized models are extracted from large, general-purpose `foundation models'  for deployment in environments with limited resources or in standalone computer systems. Although distillation uses a teacher model, it still relies on a dataset that is often limited in size and may lack representativeness or exhibit spurious correlations. In this paper, we evaluate established distillation methods, as well as the recent SubDistill method, when using data with spurious correlations for distillation. As the strength of the correlations increases, we observe a widening gap between advanced methods, such as SubDistill, which remain fairly robust, and some baseline methods, which degrade to near-random performance. Overall, our study underscores the challenges of knowledge distillation when applied to imperfect, real-world datasets, particularly those with spurious correlations.
\end{abstract}

\section{Introduction}

Recent years have seen a surge in the availability of large, general-purpose, `open-weight' ML models \cite{DBLP:journals/corr/Bommasani21,DBLP:conf/icml/RadfordKHRGASAM21}. These models can be applied off-the-shelf for tasks such as object detection and question answering, and also provide abstract representations that serve as a starting point for learning specialized tasks. While these models have shown high predictive performance, they are often too large to run on standard computers. Instead, they require expensive `inference servers', which limits their scope of application. Distillation, specifically `subtask distillation' \cite{DBLP:conf/icml/LiangZZHCZ23,shi2025selkd,chormai2026distillinglightweightdomainexperts}, offers a solution to transform these large models into more compact ones that can run on smaller machines, such as laptops or mobile phones. Recently, SubDistill \cite{chormai2026distillinglightweightdomainexperts}, a method that identifies relevant structures in the teacher's model, was proposed as an effective technique for subtask distillation.

\smallskip

In addition to developing new distillation mechanisms, it is crucial to test the performance and robustness of the proposed approaches in a wide range of realistic settings. Thus far, benchmark evaluations have primarily focused on scenarios in which the data available for distillation is a finite, yet representative, subset of the underlying data distribution. In this paper, we focus on evaluating distillation performance when the available data is affected by spurious correlations. Spurious correlations are common in many real-world datasets \cite{Calude2016}, including clinical datasets, where they arise e.g.\ from incomplete population coverage or technical confounders \cite{Zeng2022,Vaidya2024,DBLP:journals/corr/abs-2507-17845}). These correlations have been shown to give rise to `Clever Hans' classifiers \cite{lapuschkin-ncomm19, DBLP:journals/natmi/GeirhosJMZBBW20}, which exhibit unreliable predictive behavior and pose significant practical risks.

\smallskip

We conduct a benchmark evaluation of distillation techniques on popular convolutional and transformer vision models while varying the level of spurious correlation in the data. Our results demonstrate that spurious correlations significantly impact the performance of different distillation methods, albeit to varying degrees. While simple distillation baselines degrade to near-random predictions in some cases, SubDistill \cite{chormai2026distillinglightweightdomainexperts} remains fairly stable with increased levels of spurious correlations. We attribute SubDistill's stability to its superior student-teacher alignment capabilities, which prevent Clever Hans strategies from emerging. Our study also concurs with the earlier findings of \cite{bassi2024explanationneeddistillationmitigating, DBLP:conf/eccv/ParchamiAraghiBRS24}, which underscore the importance of aligning the student's and the teacher's prediction strategies. We extend their work by focusing on subtask distillation and providing a benchmark evaluation covering both recent and established distillation techniques and a variety of teacher/student architectures. Our code will be made available upon acceptance.

\section{Background}

In this section, we give a concise tutorial-focused discussion of the relevant family of distillation methods. While we emphasize simple mathematical formulations for clarity, we also provide references to corresponding methods from the literature. 

The simplest form of model distillation consists of incentivizing the student model to match the teacher's outputs through application of a loss function penalizing the difference between the two. Denoting $f_S,f_T\colon \mathbb{R}^{d} \to \mathbb{R}^C$ the functions implemented by the student and teacher models respectively, the distillation loss can be expressed as:
\begin{align}
\mathcal{E}_\text{distill}(x) = \ell(f_S(x),f_T(x)),
\label{eq:distill}
\end{align}
where $\ell$ can be the KL divergence between the student and the teacher for classification tasks \cite{DBLP:journals/corr/HintonVD15}, or the squared norm \cite{DBLP:conf/nips/BaC14} for multivariate regression. In practice, not all $C$ output dimensions of the function $f_T$ may be relevant for distillation. For example, among all image categories a foundation model can predict, only classifying between subtypes of e.g.\ `wading birds' might be relevant for the downstream application. In such a case, we can design a function $g \colon \mathbb{R}^C \to \mathbb{R}^{C'}$ that only retains the $C' \ll C$ task-relevant logits. In that case, after adapting the student model output accordingly, we can reformulate the distillation loss as:
\begin{align}
\mathcal{E}_\text{subtask}(x) = \ell(f_S(x),g \circ f_T(x)).
\label{eq:subtask}
\end{align}
To foster tighter teacher-student alignment, it is beneficial to also constrain the student's internal representations. The approach is known as layer-wise distillation and formulates a loss function of the type:
\begin{align}
\mathcal{E}_\text{layerwise}(x) = \mathcal{E}_\text{distill}(x) + \sum_{l=1}^L \lambda_l \underbrace{ \|a_S^{(l)} - a_T^{(l)}\|^2}_{\displaystyle \mathcal{E}_l(x)}
\label{eq:layerwise}
\end{align}
where $\sum_l$ iterates over a list of layer pairings between the teacher and the student, and where 
$a_S^{(l)} \in \mathbb{R}^{K_l}$ and $a_T^{(l)} \in \mathbb{R}^{d_l}$ are the student's and teacher's $l$th-layer activations associated to the data point $x$.
Because student's activation vectors typically differ in size and structure \cite{DBLP:conf/emnlp/JiaoYSJCL0L20,gu2024minillm}, the rudimentary formulation of Eq.\ \eqref{eq:layerwise} is typically enhanced with subnetworks (e.g.\ a linear map \cite{DBLP:journals/corr/RomeroBKCGB14,DBLP:conf/emnlp/JiaoYSJCL0L20} or a small MLP \cite{DBLP:conf/cvpr/AhnHDLD19}) at each layer that adapt the teacher and student representation, in particular, addressing their dimensionality mismatch.

\subsection{SubDistill Method}
\label{section:subdistill}

SubDistill is a recently proposed approach \cite{chormai2026distillinglightweightdomainexperts} that combines some of the design choices above and embeds them in a practical, numerically efficient algorithm for subtask distillation. SubDistill has two components: First, it identifies part of the representation at each layer that is most relevant for the subtask. If we consider an activation vector $a_T$ at some layer, we can construct a corresponding vector $c_T$ of the same dimensions that measures the sensitivity of those activations to the subtask of interest (details in \cite{chormai2026distillinglightweightdomainexperts}). While the product $a_T \odot c_T$ readily provides a feature-wise indicator of subtask relevance, SubDistill proceeds further by identifying a maximally relevant subspace $U$ in the teacher's activation space:
\begin{align}
    \max_U ~ \big\{\E\big[ \langle a_T, c_T \rangle_U\big] + \Omega(a_T, c_T) \big\}, 
\label{eq:prca}
\end{align} 
where we also enforce the orthogonality constraint $U^\top U= I_K$ and where $\Omega(a_T, c_T)$ stands for additional terms that ensure the objective is positive definite (cf.\ \cite{chormai2026distillinglightweightdomainexperts} for the exact objective formulation). Once teacher subspaces are identified at each layer, SubDistill constructs the following loss function for aligning teacher and student at the given layer:
\begin{align}
  \Err_l(x) =  \mathbb{E}  \big [  \big \| 
     V(a_S - \mu_S) - U^\top (a_T - \mu_T)
    \big \|^2_2   \big ]
    \label{eq:subdistill}
    \end{align}
where $\mu_S$ and $\mu_T$ are the student's and teacher's mean activations and where $V$ is constrained to be orthogonal (similar to \cite{Miles_2024_CVPR}) and represents the student's subspace rotation. The mean corrections and orthogonality constraint ensure numerical efficiency and representation alignment both before and after subspace projection. To handle potential magnitude differences, each $\Err_l$ is normalized by $\| U^\top (a_T - \mu_T) \|_2^2$.

Overall, SubDistill operationalizes the conceptual formulations in Eqs.\ \eqref{eq:distill}--\eqref{eq:layerwise} into a numerically stable and subtask-informed distillation technique.

\section{Empirical Evaluation}

\begin{table*}[t!]
        \caption{Test set accuracy of students trained with different distillation approaches under different MNIST-digit contamination rates (i.e.\ different levels of spurious correlation) on the ImageNet `wading bird' subtask. The `Ref' column is the accuracy of the teacher. The accuracy of each configuration is averaged over three random initializations, and the entry is highlighted with red, orange, and yellow if the accuracy falls in the ranges $[0, 50]$, $[50, 75]$,  and $[75, 90]$. The last column is the largest standard error of each row.}
    \label{tab:main}
    \centering
    \renewcommand{\arraystretch}{1.2}
    \begin{tabular}{cc
    >{\centering}p{1.2cm}
    >{\centering}p{1.2cm}
    >{\centering}p{1.2cm}
    >{\centering}p{1.2cm}
    >{\centering}p{1.2cm}
    >{\centering}p{1.2cm}c}
    \toprule
         Teacher $\to$ Student 
            & \makecell{Contamination\\Rate (\%)}
            & \textit{Ref.}
            & \makecell{Output\\Only\\\cite{DBLP:journals/corr/HintonVD15}}
            & \makecell{AT\\\cite{DBLP:conf/iclr/ZagoruykoK17}}
            & \makecell{VID\\\cite{DBLP:conf/cvpr/AhnHDLD19}}
            & \makecell{VKD\\\cite{Miles_2024_CVPR}}
            & \makecell{SubDistill\\\cite{chormai2026distillinglightweightdomainexperts}
            }
            & \textit{max.\ err.}
    \\
    \midrule
    ResNet18 $\to$ ResNet18-S & 0 & \textit{98.0} & 90.8 & 91.3 & 92.3 & 92.0 & 93.6 & $\pm$ 1.1 \\
     & 50 & \textit{97.6} & \cellcolor{Yellow!25} 83.9 & \cellcolor{Yellow!25} 86.9 & 91.6 & \cellcolor{Yellow!25} 89.1 & 91.7 & $\pm$ 0.9 \\
     & 100 & \textit{97.2} & \cellcolor{Red!25} 45.6 & \cellcolor{Orange!25} 52.0 & \cellcolor{Orange!25} 61.6 & \cellcolor{Orange!25} 57.2 & \cellcolor{Yellow!25} 86.4 & $\pm$ 2.1 \\
    \midrule
    WideResNet101 $\to$ MBNetv4  & 0 & \textit{98.4} & \cellcolor{Yellow!25} 90.0 & 91.3 & 90.5 & 91.1 & 96.8 & $\pm$ 1.9 \\
     & 50 & \textit{98.0} & \cellcolor{Orange!25} 62.7 & \cellcolor{Orange!25} 63.3 & \cellcolor{Orange!25} 67.1 & \cellcolor{Yellow!25} 75.1 & 92.7 & $\pm$ 1.2 \\
     & 100 & \textit{98.4} & \cellcolor{Red!25} 32.0 & \cellcolor{Red!25} 29.3 & \cellcolor{Red!25} 28.4 & \cellcolor{Red!25} 30.7 & \cellcolor{Orange!25} 66.9 & $\pm$ 3.9 \\
\midrule
     ViTB16 $\to$ EffFormerv2  & 0 & \textit{99.6} & 91.9 & 93.3 & 94.0 & 95.7 & 96.4 & $\pm$ 0.8 \\
     & 50 & \textit{99.6} & \cellcolor{Orange!25} 72.7 & \cellcolor{Orange!25} 73.3 & \cellcolor{Yellow!25} 76.8 & \cellcolor{Orange!25} 73.7 & 92.1 & $\pm$ 1.5 \\
     & 100 & \textit{99.6} & \cellcolor{Red!25} 24.0 & \cellcolor{Red!25} 24.3 & \cellcolor{Red!25} 28.1 & \cellcolor{Red!25} 28.5 & \cellcolor{Orange!25} 58.3 & $\pm$ 2.3 \\
    \bottomrule
    \end{tabular}
\end{table*}

\begin{figure}[h]
    \centering
    \includegraphics[width=\linewidth]{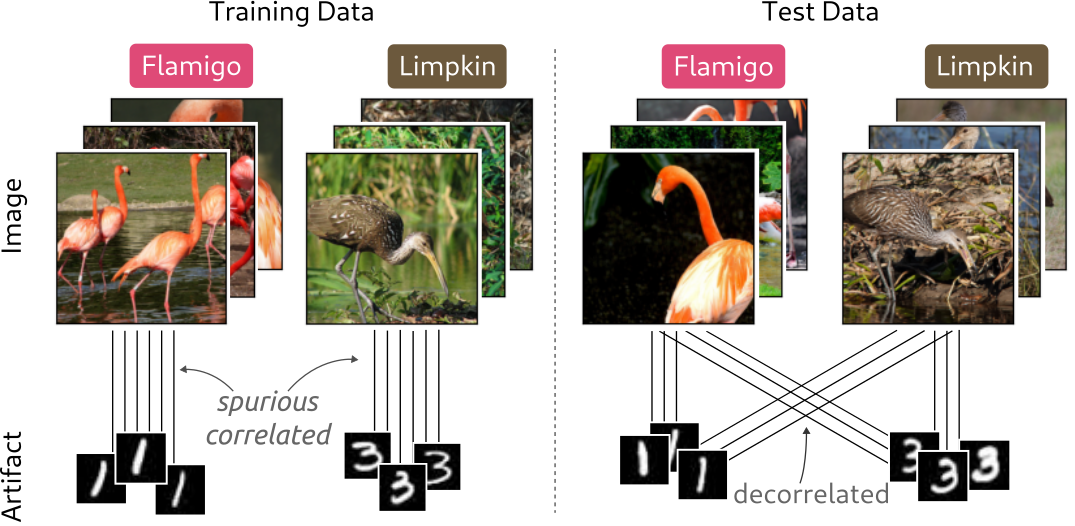}
    \caption{Spurious correlation model considered in this work, where MNIST digits are inserted in the top-right corner of ImageNet images. On the training data, the digit's class is spuriously correlated to the true image class (bird type), whereas on the test data the digits are assigned randomly.} 
    \label{fig:data}
\end{figure}

The main objective of this paper is to evaluate the performance and stability of different subtask distillation methods when trained on data exhibiting spurious correlation. For this, we consider the task of distilling from a fully trained ImageNet model a student, which we require to specialize on a specific subtask. We set as a subtask the classification of different `wading birds' (`spoonbill', `flamingo', `crane', `limpkin', and `bustard'), which form a subset of $5$ classes from the original $1000$ ImageNet classes. We synthetically add to the top left corner of some ImageNet images (of size 224$\,\times\,$224) an MNIST digit (rescaled to 56$\,\times\,$56). We generate spurious correlations in the training data by choosing MNIST digits according to the image class (i.e.\ spoonbill $\leftrightarrow$ MNIST-0, flamingo $\leftrightarrow$ MNIST-1, etc.), while we pick the digits randomly for the  test set. This model for spurious correlation is illustrated in Fig.\ \ref{fig:data}. We denote by $\rho$ the proportion of training and test ImageNet images that we contaminate and experiment with contamination rates  $\rho \in \{0\%, 50\%, 100\%\}$, which can be interpreted as adding varying levels of spurious correlation in the training data.

To cover a broad range of practical scenarios, we perform the evaluation on three teacher-student pairs. The first setup transfers from a standard ResNet18 \cite{DBLP:conf/cvpr/HeZRS16} to `ResNet18-S', a smaller ResNet18 student (for which we adjust its first three blocks and the last block to be 32 and 24 respectively). We then experiment with distilling from WideResNet101 \cite{DBLP:conf/bmvc/ZagoruykoK16} to MBNetv4 \cite{mobilenetv4}, a setup in which the teacher is substantially larger and the student has a specialized architecture aimed at deployment on mobile devices. Finally, we experiment with distilling from ViTB16 \cite{DBLP:conf/iclr/DosovitskiyB0WZ21} to EffFormerv2 \cite{li2022efficientformer}, a setup that distinguishes itself by the transformer-based architecture of these models. For each distillation experiment, we use as a teacher the corresponding ImageNet pretrained models from TorchVision \cite{torchvision2016}.

We consider in our benchmark five distillation methods ranging from the classical `output-only' method to more recent methods with enhanced student-teacher alignment, such as SubDistill. The first method, `\textit{output only}' \cite{DBLP:journals/corr/HintonVD15} induces the student and teacher to predict similarly through minimizing a KL divergence loss placed on their output probability vectors. The second method, `\textit{Attention Transfer (AT)}' \cite{DBLP:conf/iclr/ZagoruykoK17} provides finer guidance by ensuring a match between the student and teacher attention maps at each layer. `\textit{Variational Information Distillation (VID)}' \cite{DBLP:conf/cvpr/AhnHDLD19} looks at layer-wise representation more comprehensively and defines the layer-wise loss in terms of maximizing the mutual information between the teacher's and the student's representations. \textit{VKD} \cite{Miles_2024_CVPR} is a fairly recent approach that uses task-dependent normalization and orthogonal adapters to improve the alignment between the two representations. Finally, \textit{SubDistill} \cite{chormai2026distillinglightweightdomainexperts} is a recently proposed layer-wise distillation method specifically designed for subtask distillation, which we have described in Section \ref{section:subdistill}.

For all four evaluated layer-wise distillation approaches, we employ the corresponding layer-wise loss at four different teacher-student layers, similar to \cite{chormai2026distillinglightweightdomainexperts}. 
To make hyperparameter search feasible, we tie the layer-wise loss weighting coefficients $\lambda_l$ to a single parameter $\lambda$ and grid-search over  $\lambda \in \{0.01, 0.1, 1.0, 10, 100 \}$ (covering  the range used in \cite{DBLP:conf/bmvc/ZagoruykoK16,DBLP:conf/cvpr/AhnHDLD19,Miles_2024_CVPR}).
We select $\lambda$ for each method using a validation set constructed with the same contamination statistics as for the test set, similar to the setups in \cite{DBLP:conf/iclr/GulrajaniL21,pezeshki2021gradient,DBLP:conf/cvpr/0001LBYHZLZ22}.
Specifically, the validation set consists of 20\% instances held out from the original training data, and the MNIST digits we use to contaminate images are selected randomly (see  \cite{DBLP:journals/corr/abs-1907-02893,waterbird}). This can be interpreted as an `oracle' selection criteria, and each method in our benchmark benefits from it.
We train student models for 100 epochs using AdamW \cite{DBLP:conf/iclr/LoshchilovH19} and set the initial learning rate equal to $0.001$ with a decay of $0.5$ at every 25 epochs. We employ early-stopping via the accuracy on the validation set and repeat the training for each configuration for three runs (different initializations of neural network parameters and validation-set splits).

\subsection{Quantitative Results}

The performance of each tested distillation method for each distillation setup is presented in Table \ref{tab:main}. Across all considered teacher-student pairs and levels of spurious correlation, we observe the expected trend that prediction performance degrades in the presence of increased levels of spurious correlation in the training data. Another noticeable trend is that advanced methods based on layer-wise distillation, and especially the recent SubDistill method, retain higher stability under high levels of spurious correlation compared to a basic distillation scheme that only considers the model outputs. Among layer-wise distillation methods, it is noteworthy that SubDistill consistently retains accuracies above 90\% even when half of the data is contaminated with spurious correlated MNIST digits, whereas other layer-wise distillation approaches lose up to 20 percentage points. Lastly, we observe that distillation performance degrades less for the simple ResNet-18 setup compared to more complex approaches based on WideResNet/MobileNet or transformers, where some of the methods degrade to accuracy slightly above random guess (20\%) in the worst-case 100\% contamination rate.

Overall, the results presented here strongly suggest that spurious correlations need to be actively addressed through (1) data selection or generation techniques that prevent spurious correlations from occurring, (2) design of distillation techniques that enforce a tight student-teacher alignment, and (3) selection of teacher and student models for which distillation techniques are maximally effective.

\subsection{Comparing Student Representations with T-SNE}

\begin{figure*}[ht!]
    \centering
    \includegraphics[width=0.8\linewidth]{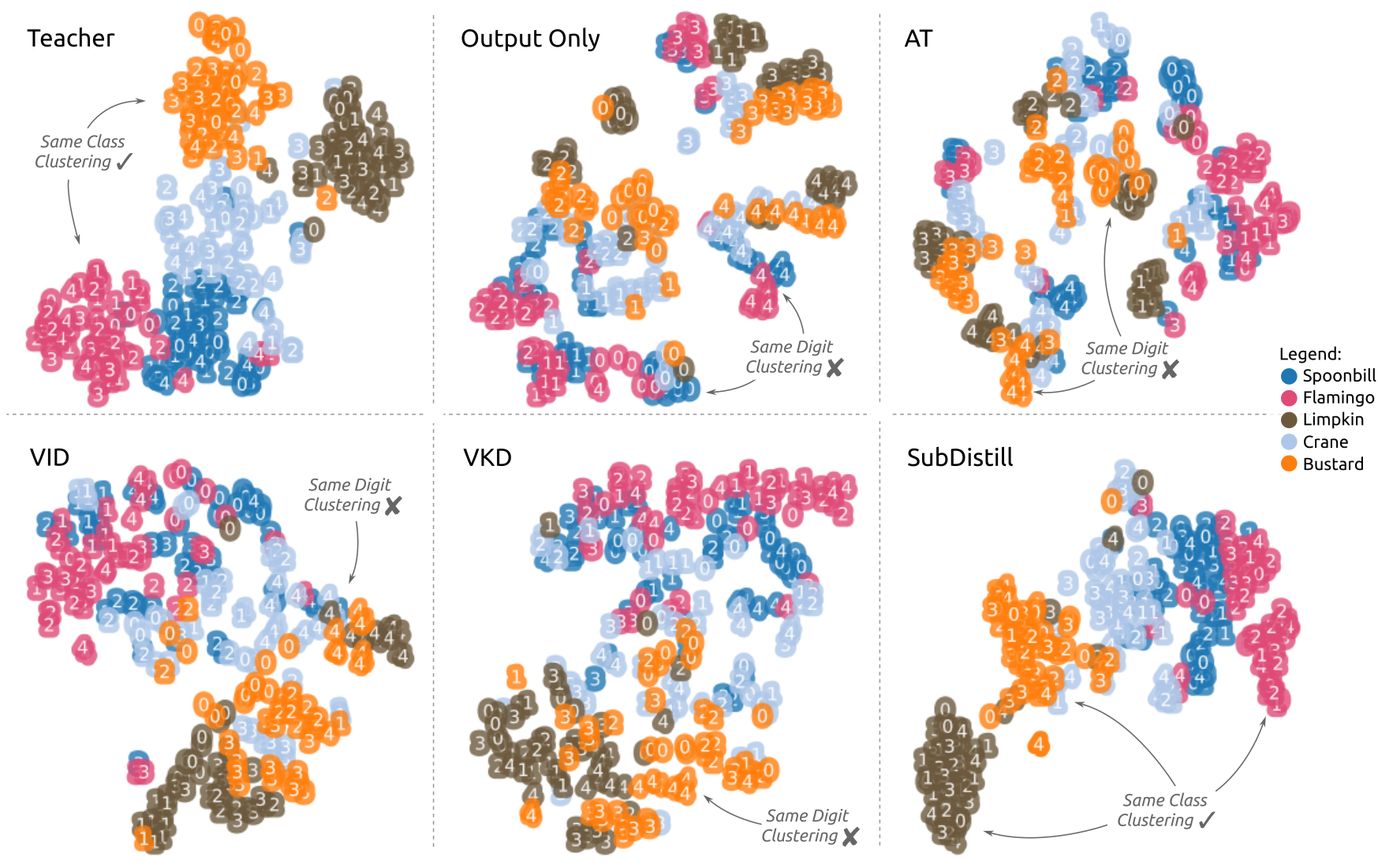}
    \caption{Data representations produced by the ResNet18 teacher and the ResNet18-S students trained with different distillation methods under the presence of spurious correlation (100\% level), and visualized using t-SNE. 
    The color of the marker corresponds to the true class label (i.e.\ the type of wading bird present in the image), while the annotated digit corresponds to the type of artifact (superposed MNIST digits 0 to 4).    
    The representations are taken from the best run of each method from the output of the global average pooling layer.}
    \label{fig:tsne}
\end{figure*}

In this section, we aim to further corroborate our findings related to the performance gap observed between distillation methods by analyzing the student models' hidden representations. Focusing on the ResNet18 pair, we investigate how semantics (of classes and spurious correlation) is encoded in the student's representation. For this purpose, we employ t-SNE \cite{JMLR:v9:vandermaaten08a} (an established non-linear dimensionality reduction algorithm) to visualize the representations of the student extracted after the global pooling layer.

Fig.~\ref{fig:tsne} shows the t-SNE visualizations of the representations from the teacher and the students trained (trained with the 100\% contamination rate, i.e.\ maximum spurious correlation).
In these visualizations, each data point is colored according to its ImageNet class, and annotated according to the class of the MNIST digit that has been added synthetically.
For the teacher, as expected, we see that its representation consists of clear clusters of colors, indicating that the class semantics rather than the MNIST artifact drive the similarity.

In contrast, the representations of the baseline students tend to form clusters that mainly code for the artifact (the MNIST digit) rather than the actual image class. This is clearly visible in the representation of the `output only' student and also visible (although to a lesser degree) for the other baseline approaches. The distortion here clearly illustrates that the spurious correlation severely affects the learning of the student, driving it to learn the semantics of the MNIST digits rather than the actual image content. This is not the case, however, for the SubDistill student, which produces a representation that distinctly resembles that of the teacher, i.e.\ with clusters that code for image content rather than the MNIST artifact. Overall, these results highlight the ability of SubDistill to ensure tight student-teacher representational alignment even in the presence of strong spurious correlations in the training data.

\begin{figure*}[t!]
    \centering
    \includegraphics[width=0.95\linewidth]{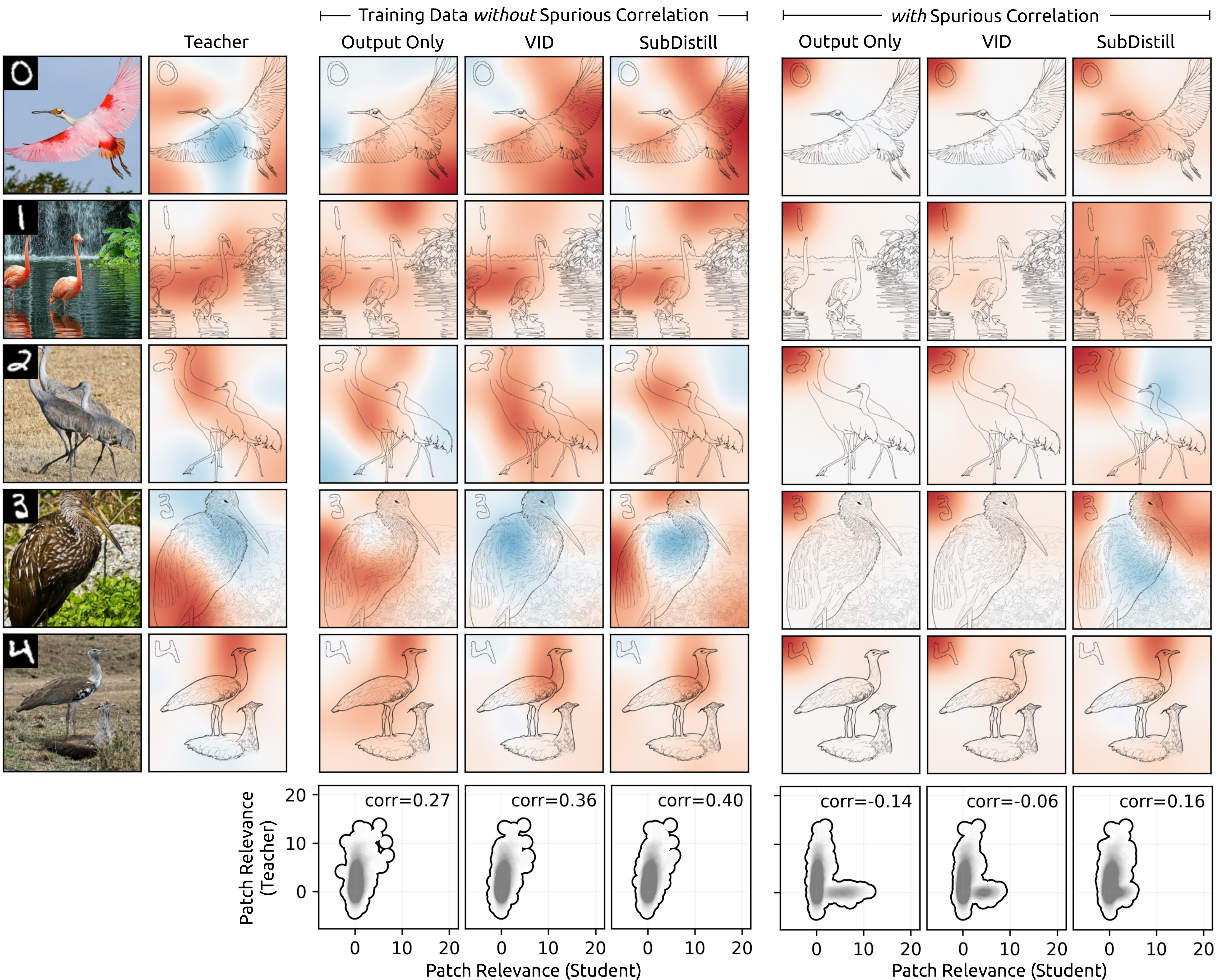}
    \caption{Top: Explainable AI analysis of the teacher and students trained with the Output Only, VID, and SubDistill approaches using the 0\%-MNIST-contaminated training data (no spurious correlation) and the 100\%-MNIST-contaminated training data (spurious correlation).
    Red highlighting indicates visual features that contribute positively to the prediction, and blue highlighting indicates features that contribute negatively.
    Bottom: Scatter plots between each patch's relevance obtained from the teacher and the student at the level of 56$\,\times\,$56 patches. The relevance scores are computed from all MNIST-contaminated test samples, averaged over the three training runs of each student.
    The depicted `corr' is the Pearson correlation coefficient.}
    \label{fig:heatmap}
\end{figure*}

\subsection{Comparing Student's Decision Strategies with XAI}

As a further qualitative comparison of the behavior of different distillation methods under spurious correlation, we use Explainable AI (XAI) \cite{DBLP:journals/aim/GunningA19,DBLP:journals/pieee/SamekMLAM21} to identify visual patterns used by each student model for predicting. Because our analysis is mainly aimed at determining whether the students focus on the signal (the image) or the top-right corner MNIST artifact, we find it sufficient and robust to apply an occlusion-based method \cite{DBLP:conf/eccv/ZeilerF14,DBLP:journals/jmlr/CovertLL21} with appropriate patch size and patch replacement scheme. We view each input image as a 4$\,\times\,$4 grid of patches of size 56$\,\times\,$56 and for each prediction we estimate the effect of removing each patch on the model output. Specifically, denoting by $x$ the input image, $x_p$ the $p$th patch, and $x_{\backslash p}$ the original image stripped from its $p$th patch, and by $f_S$ the student model, we can measure the relevance of the $p$th patch via the simple formula: 
$$
R_p(x) = f_S(x) - f_S(x_{\backslash p}).
$$
To generate the instances $x_{\backslash p}$, we make a distinction between the top-left patch containing the MNIST digit and the remaining patches forming the actual image. For the MNIST patch, we replace it with another MNIST digit of a different class. For the other patches, we simply set them to ImageNet's mean RGB. The collection of the $\{ R_p(x) \}$ constitutes the explanation and can be visualized as a heatmap of the same size as the input image.

Fig.~\ref{fig:heatmap} (top) presents the results of the XAI analysis for the cases where the students are trained with MNIST contamination rates of 0\% (clean) and 100\%. 
For the 0\% case (no spurious correlation), we see that the heatmaps of VID and SubDistill students tend to be more similar to those of the teacher than those of the output-only student. In contrast, for the 100\% case (maximum spurious correlation), we observe stronger qualitative differences, with only SubDistill showing significant robustness and ability to concentrate on the bird features, whereas students resulting from `output only' and VID primarily focus on the top-left corner (i.e.\ the artifact). The same effect is shown quantitatively in Fig.~\ref{fig:heatmap} (bottom), where we compare patch-wise relevance of the students and of the teacher over a representative set of contaminated images in the form of a scatter plot. We observe that training with spurious correlation leads to lower correlation of the teacher's and the students' relevances, becoming in the worst case negatively correlated. In this analysis again, SubDistill fares better than other approaches, and we can observe for the spurious correlation case that it is the only method that produces decision strategies that positively correlate with those of the teacher.

\section{Discussion and Conclusion}

In this work, we have investigated the emerging subtask distillation paradigm, especially whether it lends itself well to settings where the data available for training exhibits spurious correlations.

Through systematic experiments across several model pairs, including convolution and transformer architectures, we have demonstrated wide performance gaps between simple distillation baselines, which degrade to nearly random on test data, and more advanced distillation methods, specifically SubDistill, whose performance remains high. Our experiments were complemented by t-SNE and Explainable AI analyses, allowing us to link the substantial performance gap between methods to preferential encoding of signal or artifact in internal representation and to the individual models' prediction strategies.

Our work has several limitations. First, our investigation is limited in scope to the image modality and to artificially introduced spurious correlations. While such a semi-artificial setting provides fine control on the spurious correlation strength, we consider it as future work to extend the investigation to other datasets and modalities with natural forms of spurious correlations. As a further limitation, our benchmark so far focuses on existing distillation methods that are not explicitly designed to mitigate spurious correlations. We anticipate that extending distillation techniques with, e.g.\ user-in-the-loop mechanisms \cite{DBLP:journals/inffus/AndersWNSML22,cfkd-iccv}, could lead to further robustness improvements in these challenging data settings.

Overall, our work highlights significant differences in the robustness of existing distillation methods when trained under spurious correlations. Ensuring a tight student-teacher alignment in terms of both prediction and overall prediction strategy for the subtask of interest appears crucial in order to limit the potential negative impact of unrepresentative, spuriously correlated training data on the learned student model.

\section*{Acknowledgement}
This work was in part supported by the German Federal Ministry of Research, Technology and Space (BMFTR) under Grants BIFOLD24B, BIFOLD25B, 01IS18037A, 01IS18025A, and 01IS24087C.  P.C.\ is supported by the Konrad Zuse School of Excellence in Learning and Intelligent Systems (ELIZA) through the DAAD programme Konrad Zuse Schools of Excellence in Artificial Intelligence, sponsored by the Federal Ministry of Education and Research. K.R.M.\ was partly supported by the Institute of Information \& Communications Technology Planning \& Evaluation (IITP) grants funded by the Korea government (MSIT) (No. 2019-0-00079, Artificial Intelligence Graduate School Program, Korea University and No. 2022-0-00984, Development of Artificial Intelligence Technology for Personalized Plug-and-Play Explanation and Verification of Explanation). 
We thank Ali Hashemi for helpful and fruitful discussions.
\bibliographystyle{IEEEtran}
\bibliography{references}

\end{document}